# Unpaired Image Captioning by Language Pivoting

Jiuxiang Gu[1], Shafiq Joty[2], Jianfei Cai[2], Gang Wang[3]

[1] ROSE Lab, Nanyang Technological University, Singapore
[2] SCSE, Nanyang Technological University, Singapore
[3] Alibaba AI Labs, Hangzhou, China
{jgu004, srjoty, asjfcai}@ntu.edu.sg, gangwang6@gmail.com

**Abstract.** Image captioning is a multimodal task involving computer vision and natural language processing, where the goal is to learn a mapping from the image to its natural language description. In general, the mapping function is learned from a training set of image-caption pairs. However, for some language, large scale image-caption paired corpus might not be available. We present an approach to this unpaired image captioning problem by language pivoting. Our method can effectively capture the characteristics of an image captioner from the pivot language (Chinese) and align it to the target language (English) using another pivot-target (Chinese-English) sentence parallel corpus. We evaluate our method on two image-to-English benchmark datasets: MSCOCO and Flickr30K. Quantitative comparisons against several baseline approaches demonstrate the effectiveness of our method.

## 1 Introduction

Recent several years have witnessed unprecedented advancements in automatic image caption generation. This progress can be attributed *(i)* to the invention of novel deep learning framework that learns to generate natural language descriptions of images in an end-to-end fashion, and *(ii)* to the availability of large annotated corpora of images paired with captions such as MSCOCO [32] to train these models. The dominant methods are based on an encoder-decoder framework, which uses a deep convolutional neural network (CNN) to encode the image into a feature vector, and then use a recurrent neural network (RNN) to generate the caption from the encoded vector [31, 37, 29, 47]. More recently, approaches of using attention mechanisms and reinforcement learning have dominated the MSCOCO captioning leaderboard [2, 42, 20].

Despite the impressive results achieved by the deep learning framework, one performance bottleneck is the availability of large paired datasets because neural image captioning models are generally *annotation-hungry* requiring a large amount of annotated image-caption pairs to achieve effective results [21]. However, in many applications and languages, such large-scale annotations are not readily available, and are expensive and slow to acquire. In these scenarios, unsupervised methods that can generate captions from unpaired data or semi-supervised methods that can exploit paired annotations from other domains or languages are highly desirable [6]. In this paper, we pursue the later research avenue, where we assume that we have access to image-caption paired instances in one language (Chinese), and our goal is to transfer this knowledge to a target language (English) for which we do not have such image-caption paired datasets. We also assume that we have access to a separate source-target (Chinese-English) parallel corpus to help us with the transformation. In other words, we wish to use the source language (Chinese) as a pivot language to bridge the gap between an input image and a caption in the target language (English).

The concept of using a pivot language as an intermediary language has been studied previously in machine translation (MT) to translate between a resource-rich language and a resource-scarce language [50, 45, 27, 8]. The translation task in this strategy is performed in two steps. A source-to-pivot MT system first translates a source sentence into the pivot language, which is in turn translated to the target language using a pivot-to-target MT system. Although related, image captioning with the help of a pivot language is fundamentally different from MT, since it involves putting together two different tasks – captioning and translation. In addition, the pivot-based *pipelined* approach to MT suffers from two major problems when it comes to image captioning. First, the conventional pivot-based MT methods assume that the datasets for source-to-pivot and pivot-to-target translations come from the same (or similar) domain(s) with similar styles and word distributions. However, as it comes to image captioning, captions in the pivot language (Chinese) and sentences in the (Chinese-English) parallel corpus are quite different in styles and word distributions. For instance, MSCOCO captioning dataset mostly consists of images of a large scene with object instances (nouns), whereas language parallel corpora are more generic. Second, the errors made in the source-to-pivot translation get propagated to the pivot-to-target translation module in the pipelined approach.

In this paper, we present an approach that can effectively capture the characteristics of an image captioner from the source language and align it to the target language using another source-target parallel corpus. More specifically, our pivot-based image captioning framework comprises an image captioner *image-to-pivot*, an encoder-decoder model that learns to describe images in the pivot language, and a *pivot-to-target* translation model, another encoder-decoder model that translates the sentence in pivot language to the target language, and these two models are trained on two separate datasets. We tackle the variations in writing styles and word distributions in the two datasets by adapting the language translation model to the captioning task. This is achieved by adapting both the encoder and the decoder of the pivot-to-target translation model. In particular, we regularize the word embeddings of the encoder (of pivot language) and the decoder (of target language) models to make them similar to image captions. We also introduce a joint training algorithm to connect the two models and enable them to interact with each other during training. We use AIC-ICC [51] and AIC-MT [51] as the training datasets and two datasets (MSCOCO and Flickr30K [40]) as the validation datasets. The results show that our approach yields substantial gains over the baseline methods on the validation datasets.

## 2 Background

### 2.1 Image Caption Generation

Image caption generation is a fundamental problem of automatically generating natural language descriptions of images. Motivated by the recent advances in deep neural networks [22, 15] and the release of large scale datasets [32, 40, 51], many studies [48, 17, 37, 52, 24, 26, 56, 20, 19, 54] have used neural networks to generate image descriptions. Inspired by the success of encoder-decoder framework for neural machine translation (NMT) [11, 2], many researchers have proposed to use such a framework for image caption generation [47, 37]. One representative work in this direction is the method proposed by Vinyals *et al* [47]. They encode the image with a CNN and use a Long Short-Term Memory (LSTM) network as the decoder, and the decoder is trained to maximize the log-likelihood estimation of the target captions. After that, many approaches have been proposed to improve such encoder-decoder framework. One of the most commonly used approaches is the attention mechanism [53, 20]. Xu *et al* [53] use the attention mechanisms to incorporate the spatial attention on convolutional features of an image into decoder. Another improvement is to leverage the high-level visual attributes to enhance the sentence decoder [56, 55, 33]. Recently, Gu *et al* [21] propose a CNN-based image captioning model, which can explore both long-term and temporal information in word sequences for caption generation.

Exposure bias and loss-evaluation mismatch have been the major problems in sequence prediction tasks [41]. Exposure bias happens when a model is trained to predict a word given the previous ground-truth words but uses its own generated words during inference. The schedule sampling approach proposed in [3] can mitigate the exposure bias by selecting between the ground-truth words and the machine generated words according to the scheduled probability in training. Recently, the loss-evaluation mismatch problem has been well-addressed in sequence prediction tasks [41, 42, 34, 1, 20]. Rennie *et al* [42] address both exposure bias and loss-evaluation problems with a self-critical learning, which utilizes the inference mode as the baseline in training. Gu *et al* [20] propose a coarse-to-fine learning approach which simultaneously solves the multi-stage training problem as well as the exposure bias issue. The most closely related to our approach is [24]. However, they construct a multilingual parallel dataset based on MSCOCO image corpus, while in our paper, we do not have such a multilingual corpus.

### 2.2 Neural Machine Translation

Neural machine translation is an approach that directly models the conditional probability of translating a sentence in source language into a sentence in target language. A natural choice to model such a decomposition is to use RNN-based models [28, 43, 2, 36, 25, 7]. Recently, researchers have tried to improve the translation performance by introducing the attention mechanism [28, 44, 38, 12]. The attention-based translation model proposed by Kalchbrenner *et al* [28] is an early attempt to train the end-to-end NMT model. Luong *et al* [35] extend the basic encoder-decoder framework to multiple encoders and decoders. However, large-scale parallel corpora are usually not easy to obtain for some language pairs. This is unfortunate because NMT usually needs a large amount of data to train. As a result, improving NMT on resource-scarce language pairs has attracted much attention [59, 18].

Recently, many works have been done in the area of pivot strategies of NMT [13, 50, 45, 4, 57, 16, 27]. Pivot-based approach introduces a third language, named pivot language for which there exist source-pivot and pivot-target parallel corpora. The translation of pivot-based approaches can be divided into two steps: the sentence in the source language is first translated into a sentence in the pivot language, which is then translated to a sentence in the target language. However, such pivot-based approach has a major problem that the errors made in the source-to-pivot model will be forwarded to the pivot-to-target model. Recently, Cheng *et al* [9] introduce an autoencoder to reconstruct monolingual corpora. They further improve it in [10], in which they propose a joint training approach for pivot-based NMT.

## 3 Unpaired Image Captioning

Let $D_{i,x} = \{(i,x)^{(n_i)}\}_{n_i=0}^{N_i-1}$ denote the dataset with $N_i$ image-caption pairs, and $D_{x,y} = \{(x,y)^{(n_x)}\}_{n_x=0}^{N_x-1}$ denote the translation dataset with $N_x$ source-target sentence pairs. For notational simplicity, we use $i$ to denote an image instance as well as the image modality. Similarly, we use $x$ to represent a source sentence as well as a source/pivot language (Chinese), and $y$ to represent a target sentence and a target language (English). Our ultimate goal is to learn a mapping function to describe an image $i$ with a caption $y$. Formally,

$$y \sim \arg\max_{y} \left\{ P(y|i;\theta_{i\to y}) \right\} \tag{1}$$

where $\theta_{i\to y}$ are the model parameters to be learned in the absence of any paired data, $i^{(n_i)} \not\leftrightarrow y^{(n_y)}$. We use the pivot language $x$ to learn the mapping: $i \xrightarrow{\theta_{i\to x}} x \xrightarrow{\theta_{x\to y}} y$. Note that image-to-pivot ($D_{i,x}$) and pivot-to-target ($D_{x,y}$) in our setting are two distinct datasets with possibly no common elements.

Fig. 1 illustrates our pivot-based image captioning approach. We have an image captioning model $P(x|i;\theta_{i\to x})$ to generate a caption in the pivot language from an image and a NMT model $P(y|x;\theta_{x\to y})$ to translate this caption into the target language. In addition, we have an autoencoder in the target language $P(\hat{y}|\hat{y};\theta_{\hat{y}\to\hat{y}})$ that guides the target language decoder to produce caption-like sentences. We train these components jointly so that they interact with each

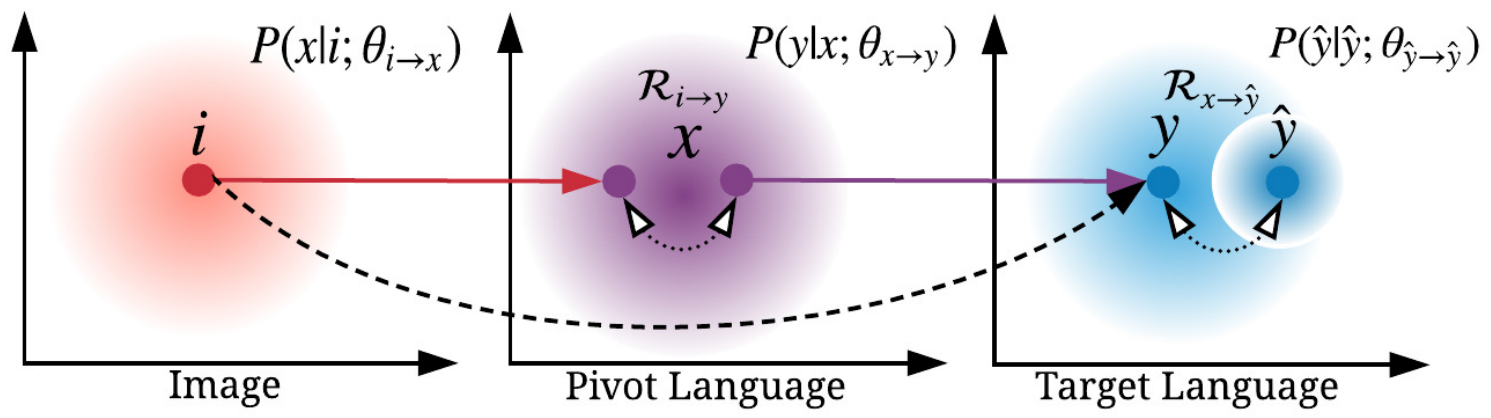

**Fig. 1.** Pictorial depiction of our pivot-based unpaired image captioning setting. Here, $i$, $x$, $y$, and $\hat{y}$ denote source image, pivot language sentence, target language sentence, and ground truth captions in target language, respectively. We use a dashed line to denote that there is no parallel corpus available for the pair. Solid lines with arrows represent decoding directions. Dashed lines inside a language (circle) denote stylistic and distributional differences between caption and translation data.

other. During inference, given an unseen image $i$ to be described, we use the joint decoder:

$$y \sim \arg\max_{y} \left\{ P(y|i; \theta_{i\to x}, \theta_{x\to y}) \right\} \tag{2}$$

In the following, we first give an overview of neural methods for image captioning and machine translation using paired (parallel) data. Then, we present our approach that extends these standard models for unpaired image captioning with a pivot language.

### 3.1 Encoder-Decoder Models for Image Captioning and Machine Translation

**Standard Image Captioning.** For image captioning in the paired setting, the goal is to generate a caption $\tilde{x}$ from an image $i$ such that $\tilde{x}$ is as similar to the ground truth caption $x$. We use $P_x(x|i; \theta_{i\to x})$ to denote a standard encoder-decoder based image captioning model with $\theta_{\mathrm{i\to x}}$ being the parameters. We first encode the given image to the image features $v$ with a CNN-based image encoder: $v = \mathrm{CNN}(i)$. Then, we predict the image description $x$ from the global image feature $v$. The training objective is to maximize the probability of the ground truth caption words given the image:

$$\tilde{\theta}_{i\to x} = \arg\max_{\theta_{i\to x}} \left\{ \mathcal{L}_{i\to x} \right\} \tag{3}$$

$$= \arg\max_{\theta_{i\to x}} \Big\{ \sum_{n_i=0}^{N_i-1} \sum_{t=0}^{M^{(n_i)}-1} \log P_x(x_t^{(n_i)} | x_{0:t-1}^{(n_i)}, i^{(n_i)}; \theta_{i\to x}) \Big\} \tag{4}$$

where $N_i$ is the number of image-caption pairs, $M^{(n_i)}$ is the length of the caption $x^{(n_i)}$, $x_t$ denotes a word in the caption, and $P_x(x_t^{(n_i)} | x_{0:t-1}^{(n_i)}, i^{(n_i)})$ corresponds to the activation of the Softmax layer. The decoded word is drawn from:

$$x_t \sim \arg\max_{\mathcal{V}_{i\to x}^{x}} P(x_t | x_{0:t-1}; i) \tag{5}$$

where $\mathcal{V}_{i\to x}^{x}$ is the vocabulary of words in the image-caption dataset $D_{i,x}$.

**Neural Machine Translation.** Given a pair of source and target sentences $(x, y)$, the NMT model $P_y(y|x; \theta_{x\to y})$ computes the conditional probability:

$$P_y(y|x) = \prod_{t=0}^{N-1} P(y_t|y_{0:t-1}; x_{0:M-1}) \tag{6}$$

where $M$ and $N$ are the lengths of the source and target sentences, respectively. The maximum-likelihood training objective of the model can be expressed as:

$$\tilde{\theta}_{x\to y} = \arg\max_{\theta_{x\to y}} \{\mathcal{L}_{x\to y}\} \tag{7}$$

$$= \arg\max_{\theta_{x\to y}} \Big\{ \sum_{n_x=0}^{N_x-1} \sum_{t=0}^{N^{(n_x)}-1} \log P_y(y_t^{(n_x)}|y_{0:t-1}^{(n_x)}; x^{(n_x)}; \theta_{x\to y}) \Big\} \tag{8}$$

During inference we calculate the probability of the next symbol given the source sentence encoding and the decoded target sequence so far, and draw the word from the dictionary according to the maximum probability:

$$y_t \sim \arg\max_{\mathcal{V}_{x\to y}^{y}} P(y_t|y_{0:t-1}; x_{0:M-1}) \tag{9}$$

where $\mathcal{V}_{x\to y}^{y}$ is the vocabulary of the target language in the translation dataset $D_{x,y}$.

### 3.2 Unpaired Image Captioning by Language Pivoting

In the unpaired setting, our goal is to generate a description $y$ in the target language for an image $i$ without any pair information. We assume, there is a second language $x$ called "pivot" for which we have (separate) image-pivot and pivot-target paired datasets. The image-to-target model in the pivot-based setting can be decomposed into two sub-models by treating the pivot sentence as a latent variable:

$$P(y|i; \theta_{i\to x}, \theta_{x\to y}) = \sum_{x} P_x(x|i; \theta_{i\to x}) P_y(y|x; \theta_{x\to y}) \tag{10}$$

where $P_x(x|i; \theta_{i\to x})$ and $P_y(y|x; \theta_{x\to y})$ are the image captioning and NMT models, respectively. Due to the exponential search space in the pivot language, we approximate the captioning process with two steps. The first step translates the image $i$ into a pivot language sentence $\tilde{x}$. Then, the pivot language sentence is translated to a target language sentence $\tilde{y}$. To learn such a pivot-based model, a simple approach is to combine the two loss functions in Equations (4) and (8) as follows:

$$\mathcal{J}_{i\to x, x\to y} = \mathcal{L}_{i\to x} + \mathcal{L}_{x\to y} \tag{11}$$

During inference, the decoding decision is given by:

$$\tilde{x} = \arg\max_{x} \left\{ P_x(x|i; \tilde{\theta}_{i\to x}) \right\} \tag{12}$$

$$\tilde{y} = \arg\max_{y} \left\{ P_y(y|\tilde{x}; \tilde{\theta}_{x\to y}) \right\} \tag{13}$$

where $\tilde{x}$ is the image description generated from $i$ in the pivot language, $\tilde{y}$ is the translation of $\tilde{x}$, and $\tilde{\theta}_{i\to x}$ and $\tilde{\theta}_{x\to y}$ are the learned model parameters.

However, this *pipelined* approach to image caption generation in the target language suffers from couple of key limitations. First, image captioning and machine translation are two different tasks. The image-to-pivot and pivot-to-target models are quite different in terms of vocabulary and parameter space because they are trained on two possibly unrelated datasets. Image captions contain description of objects in a given scene, whereas machine translation data is more generic, in our case containing news event descriptions, movie subtitles, and conversational texts. They are two different domains with differences in writing styles and word distributions. As a result, the captions generated by the pipeline approach may not be similar to human-authored captions. Fig. 1 distinguishes between the two domains of pivot and target sentences: caption domain and translation domain (see second and third circles). The second limitation is that the errors made by the image-to-pivot captioning model gets propagated to the pivot-to-target translation model.

To overcome the limitations of the pivot-based caption generation, we propose to reduce the discrepancy between the image-to-pivot and pivot-to-target models, and to train them jointly so that they learn better models by interacting with each other during training. Fig. 2 illustrates our approach. The two models share some common aspects that we can exploit to connect them as we describe below.

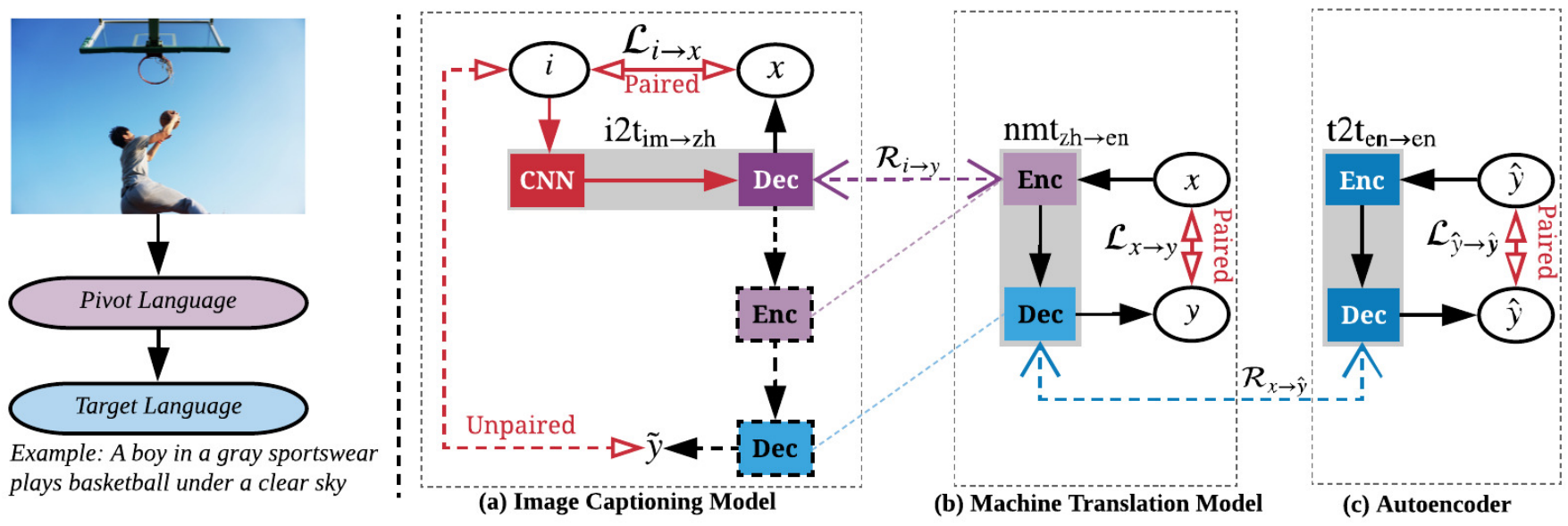

**Fig. 2.** Illustration of our image captioning model with pivot language. The image captioning model first transforms an image into latent pivot sentences, from which our machine translation model generates the target caption.

**Connecting Image-to-Pivot and Pivot-to-Target.** One way to connect the two models is to share the corresponding embedding matrices by defining a common embedding matrix for the decoder of image-to-pivot and the encoder of pivot-to-target. However, since the caption and translation domains are different, their word embeddings should also be different. Therefore, rather than having a common embedding matrix, we add a regularizer $\mathcal{R}_{i\to y}$ that attempts to bring the input embeddings of the NMT model close to the output embeddings of image captioning model by minimizing their $l_2$ distance. Formally,

$$\mathcal{R}_{i\to y}(\theta^{w_x}_{i\to x}, \theta^{w_x}_{x\to y}) = -\sum_{w_x \in \mathcal{V}^x_{i\to x} \cap \mathcal{V}^x_{x\to y}} ||\theta^{w_x}_{i\to x} - \theta^{w_x}_{x\to y}||_2 \tag{14}$$

where $w_x$ is a word in the pivot language that is shared by the two embedding matrices, and $\theta^{w_x}_{i\to x} \in \mathbb{R}^d$ denotes the vector representation of $w_x$ in the source-to-pivot model, and $\theta^{w_x}_{x\to y} \in \mathbb{R}^d$ denotes the vector representation of $w_x$ in the pivot-to-target model. Note that, here we adapt $\theta^{w_x}_{x\to y}$ towards $\theta^{w_x}_{i\to x}$, that is, $\theta^{w_x}_{i\to x}$ is already a learned model and kept fixed during adaptation.

Adapting the encoder embeddings of the NMT model does not guarantee that the decoder of the model will produce caption-like sentences. For this, we need to also adapt the decoder embeddings of the NMT model to the caption data. We first use the target-target parallel corpus $D_{\hat{y},\hat{y}} = \{(\hat{y}^{(n_{\hat{y}})}, \hat{y}^{(n_{\hat{y}})})\}_{n_{\hat{y}}=0}^{N_{\hat{y}}-1}$ to train an autoencoder $P(\hat{y}|\hat{y}; \theta_{\hat{y}\to\hat{y}})$, where $\theta_{\hat{y}\to\hat{y}}$ are the parameters of the autoencoder. The maximum-likelihood training objective of autoencoder can be expressed as:

$$\tilde{\theta}_{\hat{y}\to\hat{y}} = \arg\max_{\theta_{\hat{y}\to\hat{y}}} \left\{\mathcal{L}_{\hat{y}\to\hat{y}}\right\} \tag{15}$$

where $\mathcal{L}_{\hat{y}\to\hat{y}}$ is the cross-entropy (XE) loss. The autoencoder then "teaches" the decoder of the translation model $P(y|x; \theta_{x\to y})$ to learn similar word representations. This is again achieved by minimizing the $l_2$ distance between two vectors:

$$\mathcal{R}_{x\to\hat{y}}(\theta^{w_y}_{x\to y}, \theta^{w_y}_{\hat{y}\to\hat{y}}) = -\sum_{w_y \in \mathcal{V}^y_{x\to y} \cap \mathcal{V}^y_{\hat{y}\to\hat{y}}} ||\theta^{w_y}_{x\to y} - \theta^{w_y}_{\hat{y}\to\hat{y}}||_2 \tag{16}$$

where $\mathcal{V}^y_{\hat{y}\to\hat{y}}$ is the vocabulary of $y$ in $D_{\hat{y},\hat{y}}$, and $w_y$ is a word in the target language that is shared by the two embedding matrices. By optimizing Equation (16), we try to make the learned caption share a similar style as the target captions.

**Joint Training.** In training, our goal is to find a set of source-to-target model parameters that maximizes the training objective:

$$\mathcal{J}_{i\to x, x\to y, y\to\hat{y}} = \mathcal{L}_{i\to x} + \mathcal{L}_{x\to y} + \mathcal{L}_{\hat{y}\to\hat{y}} + \lambda \mathcal{R}_{i\to x, x\to y, y\to\hat{y}} \tag{17}$$

$$\mathcal{R}_{i\to x, x\to y, y\to\hat{y}} = \mathcal{R}_{i\to y}(\theta^{w_x}_{i\to x}, \theta^{w_x}_{x\to y}) + \mathcal{R}_{x\to\hat{y}}(\theta^{w_y}_{x\to y}, \theta^{w_y}_{\hat{y}\to\hat{y}}) \tag{18}$$

where $\lambda$ is the hyper-parameter used to balance the preference between the loss terms and the connection terms. Since both the captioner $P_x(x|i; \theta_{i\to x})$ and the translator $P_y(y|x; \theta_{x\to y})$ have large vocabulary sizes (see Table 1), it is hard to train the joint model with an initial random policy. Thus, in practice, we pre-train the captioner, translator and autoencoder first, and then jointly optimize them with Equation (17).

## 4 Experiments

**Datasets.** In our experiments, we choose the two independent datasets used from AI Challenger (AIC[4]) [51]: AIC Image Chinese Captioning (AIC-ICC) and AIC Chinese-English Machine Translation (AIC-MT), as the training datasets, while using MSCOCO and Flickr30K English captioning datasets as the test datasets. Table 1 shows the statistics of the datasets used in our experiments. Fig. 3 illustrates the differences in word-level distributions among AIC-ICC, AIC-MT, and MSCOCO using word clouds[5].

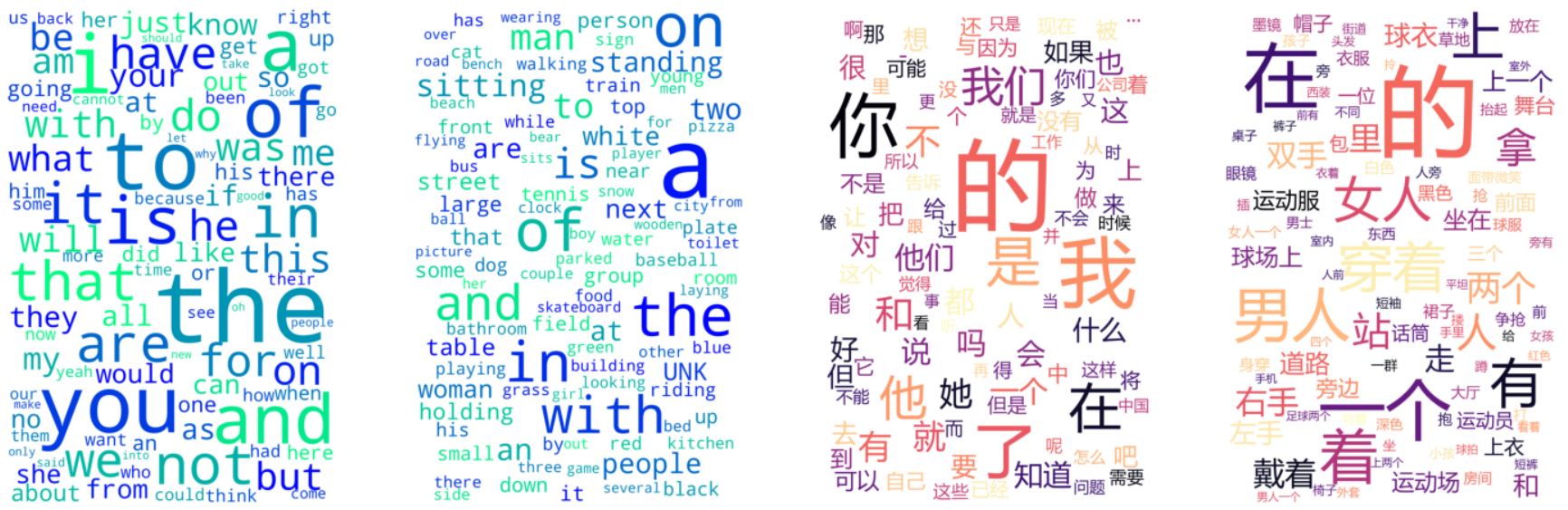

(a) AIC-MT (en) (b) MSCOCO (en) (c) AIC-MT (zh) (d) AIC-ICC (zh)

**Fig. 3.** Word clouds of AIC-MT, MSCOCO, and AIC-ICC datasets, where different font sizes indicate different frequencies of words.

**Table 1.** Statistics of the datasets used in our experiments, where "im" denotes the image, "zh" denotes Chinese, and "en" denotes English.

| | Dataset | Lang. | Source | | Target | |
|---|---|---|---|---|---|---|
| | | | # Image/Sent. | Vocab. Size | # Sent. | Vocab. Size |
| Training | AIC-ICC | im → zh | 240K | – | 1,200K | 4,461 |
| | AIC-MT | zh → en | 10,000K | 50,004 | 10,000K | 50,004 |
| Testing | MSCOCO | im → en | 123K | – | 615K | 9,487 |
| | Flickr30K | im → en | 30K | – | 150K | 7,000 |

**Training Datasets.** For image-to-Chinese captioning training, we follow the settings in AIC-ICC [51] and take 210,000 images for training and 30,000 images for model validation. Each image contains five reference Chinese captions, and each of the captions contains most of the common daily scenes in which a person usually appears (see Fig. 3(d)). We use the "*Jieba*"[6], a Chinese text segmentation module, for word segmentation. We truncate all the captions longer than 16 tokens and prune the vocabulary by dropping words with a frequency less than 5, resulting in a vocabulary size of 4,461 words.

We take Chinese as the pivot language and learn a Chinese-to-English translation model on AIC-MT. AIC-MT consists of 10,000K Chinese-English parallel sentences, and the English sentences are extracted from English learning websites and movie subtitles. We reserve 4K sentence pairs for validation and 4K sentence pairs for testing. During preprocessing, we remove the empty lines and retain sentence pairs with no more than 50 words. We prune the vocabulary and end up with a vocabulary of size 50,004 words including special Begin-of-Sentence (BOS) and End-of-Sentence (EOS) tokens. To guide the target decoder to generate caption-like sentences, an autoencoder is also trained with the target image descriptions extracted from MSCOCO. In our training, we extract 60,000 image descriptions from the MSCOCO training split, and randomly sort these samples.

[4] https://challenger.ai/competition
[5] https://pypi.python.org/pypi/wordcloud
[6] https://github.com/fxsjy/jieba

**Validation Datasets.** We validate the effectiveness our method on MSCOCO and Flickr30K datasets. The images in MSCOCO typically contain multiple objects with significant contextual information. Likewise, each image in MSCOCO also has five reference description, and most of these descriptions are depicting humans participating in various activities. We use the same test splits as that in [29]. For MSCOCO, we use 5,000 images for validation and 5,000 images for testing, and for Flickr30K, we use 1,000 images for testing.

### 4.1 Implementation Details

**Architecture.** As can be seen in Fig. 2, we have three models used in our image captioner. The first model $\text{i2t}_{\text{im}\rightarrow\text{zh}}$ learns to generate the Chinese caption $x$ from a given image $i$. It is a standard CNN-RNN architecture [49], where word outputted from the previous time step is taken as the input for the current time step. For each image, we encoder it with ResNet-101 [23], and then apply average pooling to get a vector of dimensions 2,048. After that, we map the image features through a linear projection and get a vector of dimensions 512. The decoder is implemented based on a LSTM network. The dimensions of the LSTM hidden states and word embedding are fixed to 512 for all of the models discussed in this paper. Each sentence starts with a special BOS token, and ends with an EOS token.

The second model $\text{nmt}_{\text{zh}\rightarrow\text{en}}$ learns to translate the Chinese sentence $x$ to the English sentence $y$. It has three components: a sentence encoder, a sentence decoder, and an attention module. The words in the pivot language are first mapped to word vectors and then fed into a bidirectional LSTM network. The decoder predicts the target language words based on the the encoded vector of the source sentence as well as its previous outputs. The encoder and the decoder are connected through an attention module which allows the decoder to focus on different regions of the source sentence during decoding.

The third model $\text{t2t}_{\text{en}\rightarrow\text{en}}$ learns to produce the caption-style English sentence $\hat{y}$. It is essentially an autoencoder trained on a set of image descriptions extracted from MSCOCO, where the encoder and the decoder are based on one-layer LSTM network. The encoder reads the whole sentence as input and the decoder is to reconstruct the input sentence.

**Training Setting.** All the modules are randomly initialized before training except the image CNN, for which we use a pre-trained model on ImageNet. We first independently train the image Chinese captioner, the Chinese-to-English translator, and the autoencoder with the cross-entropy loss on AIC-ICC, AIC-MT and MSCOCO corpus, respectively. During this stage, we use Adam [30] algorithm to do model updating with a mini-batch size of 100. The initial learning rate is $4e^{-4}$, and the momentum is 0.9. The best models are selected according to the validation scores, which are then used for the subsequent joint training. Specifically, we combine the just trained models with the connection terms, and conduct a joint training with (17). We set the hyper-parameter $\lambda$ to 1.0, and train the joint model using Adam optimizer with a mini-batch size of 64 and an initial learning rate of $2e^{-4}$. Weight decay and dropout are applied in this training phase to prevent over-fitting.

**Testing setting.** During testing, the output image description is first formed by drawing words in pivot language from $\text{i2t}_{\text{im}\to\text{zh}}$ until an EOS token is reached, and then translated with $\text{nmt}_{\text{zh}\to\text{en}}$ to the target language. Here we use beam search for the two inference procedures. Beam search is an efficient decoding method for RNN-based models, which keeps the top-$k$ hypotheses at each time step, and considers them as the candidates to generate a new top-$k$ hypotheses at the next time step. We set a fixed beam search size of $k = 5$ for $\text{i2t}_{\text{im}\to\text{zh}}$ and $k = 10$ for $\text{nmt}_{\text{zh}\to\text{en}}$. We evaluate the quality of the generated image descriptions with the standard evaluation metrics: BLEU [39], METEOR [14], and CIDEr [46]. Since BLEU aims to assess how similar two sentences are, we also evaluate the diversity of the generated sentence with Self-BLEU [58], which takes one sentence as the hypothesis and the others as the reference, and then calculates BLEU score for every generated sentence. The final Self-BLEU score is defined as the average BLEU scores of the sentences.

### 4.2 Quantitative Analysis

**Results of Image Chinese Captioning.** Table 2 shows the comparison results on the AIC-ICC validation set, where B@n is short for BLEU-n. We compare our $\text{i2t}_{\text{im}\to\text{zh}}$ model with the baseline [51] (named AIC-I2T). Both AIC-I2T and our image caption model ($\text{i2t}_{\text{im}\to\text{zh}}$) are trained with cross-entropy loss. We can see that our model outperforms the baseline in all the metrics. This might be due to different implementation details, e.g. AIC-I2T utilizes Inception-v3 for the image CNN while we use ResNet-101.

**Table 2.** Performance comparisons on AIC-ICC. The results of $\text{i2t}_{\text{im}\to\text{zh}}$ are achieved via beam search.

| Image Chinese Captioning | | | | | |
|---|---|---|---|---|---|
| Approach | B@1 | B@2 | B@3 | B@4 | CIDEr |
| $\text{i2t}_{\text{im}\to\text{zh}}$ | **77.8** | **65.9** | **55.5** | **46.6** | **144.2** |
| AIC-I2T [51] | 76.5 | 64.8 | 54.7 | 46.1 | 142.5 |

**Table 3.** Performance comparisons on AIC-MT test dataset. Note that our $\text{nmt}_{\text{zh}\to\text{en}}$ model uses beam search.

| Chinese-to-English Translation | | |
|---|---|---|
| Approach | Accuracy | Perplexity |
| $\text{nmt}_{\text{zh}\to\text{en}}$ | 55.0 | 8.9 |
| Google Translation | **57.8** | – |

**Results of Chinese-to-English Translation.** Table 3 provides the comparison between our attention-based machine translator with the online Google translator on AIC-MT test split. We use the "googletrans"[7], a free Python tool that provides Google translator API. The perplexity value in the second column is the geometric mean of the inverse probability for each predicted word. Our attention-based NMT model ($\text{nmt}_{\text{zh}\to\text{en}}$) is trained on AIC-MT training set. We can see that our model is slightly worse than online Google translation in accuracy. This not surprising considering that Google's translator is trained on much larger datasets with more vocabulary coverage, and it is a more complex system that ensembles multiple NMT models.

**Results of Unpaired Image English Captioning.** Table 4 shows the comparisons among different variants of our method on MSCOCO dataset. Our upper bound is achieved by an image captioning model $\text{i2t}_{\text{im}\to\text{en}}$ that is trained with paired English captions. $\text{i2t}_{\text{im}\to\text{en}}$ shares the same architecture as $\text{i2t}_{\text{im}\to\text{zh}}$, except that they have different vocabulary sizes. The lower bound is achieved by pipelinining $\text{i2t}_{\text{im}\to\text{zh}}$ and $\text{nmt}_{\text{zh}\to\text{en}}$. In the pipeline setting, these two models are trained on AIC-ICC and AIC-MT, respectively. We also report the results of our implementation of FC-2K [42], which adopts a similar architecture.

**Table 4.** Results of unpaired image-to-English captioning on MSCOCO 5K and Flickr30K 1K test splits.

| Approach | Lang. | B@1 | B@2 | B@3 | B@4 | M | CIDEr |
|---|---|---|---|---|---|---|---|
| *MSCOCO* | | | | | | | |
| $\text{i2t}_{\text{im}\to\text{en}}$ (Upper bound, XE Loss) | en | 73.2 | 56.3 | 42.0 | 31.2 | 25.3 | 95.1 |
| FC-2K [42] (ResNet101, XE Loss) | en | – | – | – | 29.6 | 25.2 | 94.0 |
| $\text{i2t}_{\text{im}\to\text{zh}}$ + $\text{nmt}_{\text{zh}\to\text{en}}$ ($\mathcal{R}_{i\to x,x\to y,y\to\hat{y}}$) | en | **46.2** | **24.0** | **11.2** | **5.4** | 13.2 | **17.7** |
| $\text{i2t}_{\text{im}\to\text{zh}}$ + $\text{nmt}_{\text{zh}\to\text{en}}$ ($\mathcal{R}_{i\to x,x\to y}$) | en | 45.5 | 23.6 | 11.0 | 5.3 | 13.1 | 17.3 |
| $\text{i2t}_{\text{im}\to\text{zh}}$ + $\text{nmt}_{\text{zh}\to\text{en}}$ (Lower bound) | en | 42.0 | 20.6 | 9.5 | 3.9 | 12.0 | 12.3 |
| $\text{i2t}_{\text{im}\to\text{zh}}$ + Online Google Translation | en | 42.2 | 21.8 | 10.7 | 5.3 | **14.5** | 17.0 |
| *Flickr30K* | | | | | | | |
| $\text{i2t}_{\text{im}\to\text{en}}$ (Upper bound, XE Loss) | en | 63.1 | 43.8 | 30.2 | 20.7 | 17.7 | 40.1 |
| $\text{i2t}_{\text{im}\to\text{zh}}$ + $\text{nmt}_{\text{zh}\to\text{en}}$ ($\mathcal{R}_{i\to x,x\to y,y\to\hat{y}}$) | en | **49.7** | **27.8** | **14.8** | **7.9** | 13.6 | **16.2** |
| $\text{i2t}_{\text{im}\to\text{zh}}$ + $\text{nmt}_{\text{zh}\to\text{en}}$ ($\mathcal{R}_{i\to x,x\to y}$) | en | 48.7 | 26.1 | 12.8 | 6.4 | 13.0 | 14.9 |
| $\text{i2t}_{\text{im}\to\text{zh}}$ + $\text{nmt}_{\text{zh}\to\text{en}}$ (Lower bound) | en | 45.9 | 25.2 | 13.1 | 6.9 | 12.5 | 13.9 |
| $\text{i2t}_{\text{im}\to\text{zh}}$ + Online Google Translation | en | 46.2 | 25.4 | 13.9 | 7.7 | **14.4** | 15.8 |

For unpaired image-to-English captioning, our method with the connection term on pivot language ($\mathcal{R}^{w_x}_{i\to x,x\to y}$) outperforms the method of combining $\text{i2t}_{\text{im}\to\text{zh}}$ with online Google translation in terms of B@n and CIDEr metrics, while obtaining significant improvements over the lower bound. This demonstrates the effectiveness of the connection term on the pivot language. Moreover, by adding the connection term on the target language, our model with the two connection terms ($\mathcal{R}_{i\to x,x\to y,y\to\hat{y}}$) further improves the performance. This suggests that a small corpus in the target domain is able to make the decoder to generate image descriptions that are more like captions. The connection terms help to bridge the word representations of the two different domains. The captions generated by Google translator have higher METEOR. We speculate the following reasons. First, Google Translator generates longer captions than ours. Since METEOR computes the score not only on the basis of $n$-gram precision but also of uni-gram recall, its default parameters favor longer translations than other metrics [5]. Second, in addition to exact word matching, METEOR considers matching of word stems and synonyms. Since Google translator is trained on a much larger corpus than ours, it generates more synonymous words. Table 4 also shows the results of unpaired image English captioning on Flickr30K, where we can draw similar conclusions.

[7] https://pypi.python.org/pypi/googletrans

We further evaluate the diversity of the generated image descriptions using Self-BLEU metric. Table 5 shows the detailed Self-BLEU scores. It can be seen that our method generates image descriptions with the highest diversity, compared with the upper and lower bounds. For better comparison, we also calculate the Self-BLEU scores calculated on ground-truth captions.

**Table 5.** Self-BLEU scores on MSCOCO 5K test split. Note that lower Self-BLEU scores imply higher diversity of the image descriptions.

| Approach | Lang. | Self-B@2 | Self-B@3 | Self-B@4 | Self-B@5 |
|---|---|---|---|---|---|
| $\text{i2t}_{\text{im}\to\text{en}}$ (GT Captions) | en | 85.0 | 67.8 | 49.1 | 34.4 |
| $\text{i2t}_{\text{im}\to\text{en}}$ (Upper bound) | en | 99.0 | 97.5 | 94.6 | 90.7 |
| $\text{i2t}_{\text{im}\to\text{zh}} + \text{nmt}_{\text{zh}\to\text{en}}$ ($\mathcal{R}_{i\to x, x\to y, y\to\hat{y}}$) | en | **95.6** | **91.7** | **86.5** | **80.2** |
| $\text{i2t}_{\text{im}\to\text{zh}} + \text{nmt}_{\text{zh}\to\text{en}}$ (Lower bound) | en | 98.1 | 95.9 | 92.3 | 87.6 |

**Table 6.** Evaluation results of user assessment on MSCOCO 1.2K test split.

| Approach | $\text{i2t}_{\text{im}\to\text{zh}} + \text{nmt}_{\text{zh}\to\text{en}}$ ($\mathcal{R}_{i\to x, x\to y, y\to\hat{y}}$) | Upper Bound | Ground-Truth |
|---|---|---|---|
| Relevant | 3.81 | 3.99 | 4.68 |
| Resemble | 3.78 | 4.05 | 4.48 |

We also conduct a human evaluation of the generated captions for different models as well as the ground truth captions. A total number of 12 evaluators of different educational background were invited, and a total of 1.2K samples were randomly selected from the test split for the user study. Particularly, we measure the caption quality from two aspects: *relevant* and *resemble*. The *relevant* metric indicates whether the caption is correct according to the image content. The *resemble* metric assesses to what extent the systems produce captions that resemble human-authored captions. The evaluators assessed the quality in 5 grades: 1-very poor, 2-poor, 3-barely acceptable, 4-good, 5-very good. Each evaluator assessed randomly chosen 100 images. The results presented in Table 6 demonstrate that our approach can generate relevant and human understandable image captions as the paired (Upper bound) approach.

### 4.3 Qualitative Results

We provide some captioning examples in Fig. 4 for a better understanding of our model. We show in different color the generated captions for several images by the three models along with the ground truth (GT) captions. From these exemplary results, we can see that, compared with the paired model $\text{i2t}_{\text{im}\rightarrow\text{en}}$, our pivot-based unpaired model $\text{i2t}_{\text{im}\rightarrow\text{zh}\rightarrow\text{en}}(\mathcal{R}_{i\rightarrow x,x\rightarrow y,y\rightarrow \hat{y}})$ often generates more diverse captions; thanks to the additional translation data. At the same time, our model can generate caption-like sentences by bridging the gap between the datasets and by joint training of the model components. For example, with the detected people in the first image, our model generates the sentence with "*a bunch of people in sports suits*", which is more diverse than the sentence with "*a group of baseball players*" generated by the paired model.

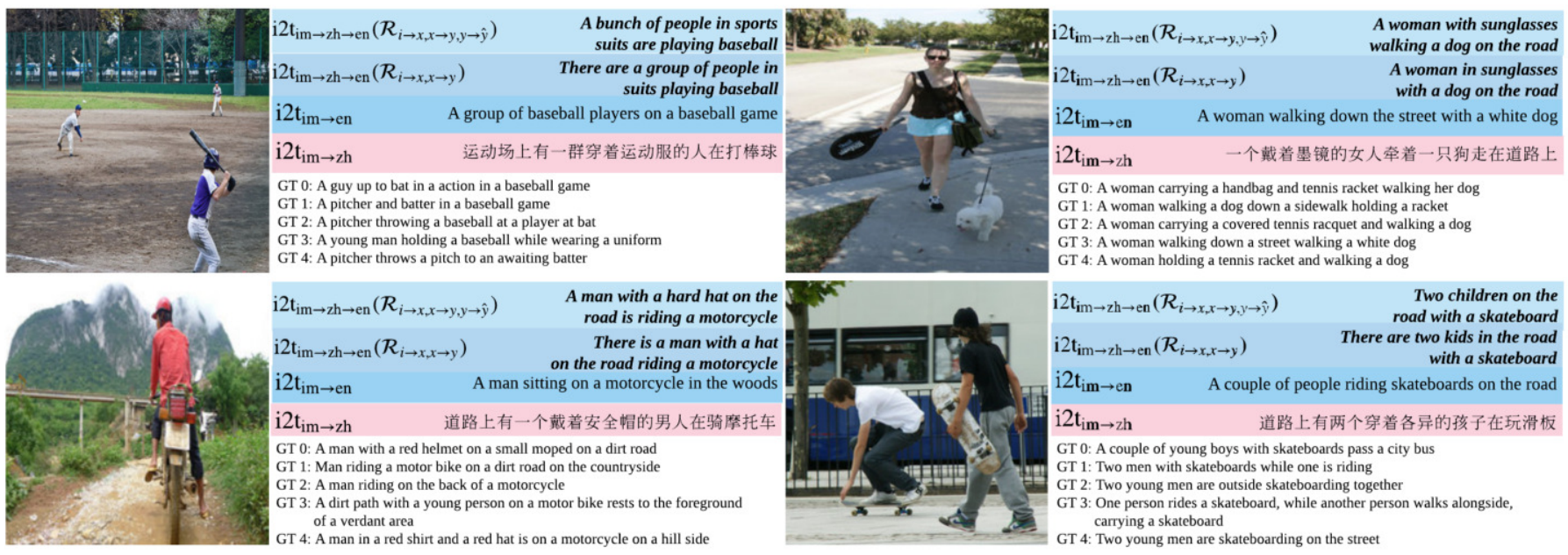

**Fig. 4.** Examples of the generated sentences on MSCOCO test images, where $\text{i2t}_{\text{im}\rightarrow\text{zh}}$ is the image captioner trained on AIC-ICC, $\text{i2t}_{\text{im}\rightarrow\text{en}}$ is the image captioner trained on MSCOCO, $\text{i2t}_{\text{im}\rightarrow\text{zh}\rightarrow\text{en}}$ $(\mathcal{R}_{i\rightarrow x,x\rightarrow y,y\rightarrow \hat{y}})$ and $\text{i2t}_{\text{im}\rightarrow\text{zh}\rightarrow\text{en}}$ $(\mathcal{R}_{i\rightarrow x,x\rightarrow y})$ are our proposed models for unpaired image captioning, and GT stands for ground-truth caption.

## 5 Conclusion

In this paper, we have proposed an approach to unpaired image captioning with the help of a pivot language. Our method couples an image-to-pivot captioning model with a pivot-to-target NMT model in a joint learning framework. The coupling is done by adapting the word representations in the encoder and the decoder of the NMT model to produce caption-like sentences. Empirical evaluation demonstrates that our method consistently outperforms the baseline methods on MSCOCO and Flickr30K image captioning datasets. In our future work, we plan to explore the idea of 'back-translation' to create pseudo Chinese-English translation data for English captions, and adapt our decoder language model by training on this pseudo dataset.

## Acknowledgments

This research was carried out at the Rapid-Rich Object Search (ROSE) Lab at the Nanyang Technological University, Singapore. The ROSE Lab is supported by the National Research Foundation, Singapore, and the Infocomm Media Development Authority, Singapore. We gratefully acknowledge the support of NVIDIA AI Tech Center (NVAITC) for our research at NTU ROSE Lab, Singapore.

## References

1. Bahdanau, D., Brakel, P., Xu, K., Goyal, A., Lowe, R., Pineau, J., Courville, A., Bengio, Y.: An actor-critic algorithm for sequence prediction. In: ICLR (2017)
2. Bahdanau, D., Cho, K., Bengio, Y.: Neural machine translation by jointly learning to align and translate. In: ICLR (2015)
3. Bengio, S., Vinyals, O., Jaitly, N., Shazeer, N.: Scheduled sampling for sequence prediction with recurrent neural networks. In: NIPS (2015)
4. Bertoldi, N., Barbaiani, M., Federico, M., Cattoni, R.: Phrase-based statistical machine translation with pivot languages. In: IWSLT (2008)
5. Cer, D., Manning, C.D., Jurafsky, D.: The best lexical metric for phrase-based statistical mt system optimization. In: ACL (2010)
6. Chen, T.H., Liao, Y.H., Chuang, C.Y., Hsu, W.T., Fu, J., Sun, M.: Show, adapt and tell: Adversarial training of cross-domain image captioner. In: ICCV (2017)
7. Chen, W., Matusov, E., Khadivi, S., Peter, J.T.: Guided alignment training for topic-aware neural machine translation. In: AMTA (2016)
8. Chen, Y., Liu, Y., Li, V.O.: Zero-resource neural machine translation with multi-agent communication game. In: AAAI (2018)
9. Cheng, Y., Xu, W., He, Z., He, W., Wu, H., Sun, M., Liu, Y.: Semi-supervised learning for neural machine translation. In: ACL (2016)
10. Cheng, Y., Yang, Q., Liu, Y., Sun, M., Xu, W.: Joint training for pivot-based neural machine translation. In: Proceedings of IJCAI (2017)
11. Cho, K., Van Merriënboer, B., Gulcehre, C., Bahdanau, D., Bougares, F., Schwenk, H., Bengio, Y.: Learning phrase representations using rnn encoder-decoder for statistical machine translation. EMNLP (2014)
12. Cohn, T., Hoang, C.D.V., Vymolova, E., Yao, K., Dyer, C., Haffari, G.: Incorporating structural alignment biases into an attentional neural translation model. In: ACL (2016)
13. Cohn, T., Lapata, M.: Machine translation by triangulation: Making effective use of multi-parallel corpora. In: ACL (2007)
14. Denkowski, M., Lavie, A.: Meteor universal: Language specific translation evaluation for any target language. In: ACL (2014)
15. Ding, H., Jiang, X., Shuai, B., Liu, A.Q., Wang, G.: Context contrasted feature and gated multi-scale aggregation for scene segmentation. In: CVPR (2018)
16. El Kholy, A., Habash, N., Leusch, G., Matusov, E., Sawaf, H.: Language independent connectivity strength features for phrase pivot statistical machine translation. In: ACL (2013)
17. Fang, H., Gupta, S., Iandola, F., Srivastava, R.K., Deng, L., Dollár, P., Gao, J., He, X., Mitchell, M., Platt, J.C., et al.: From captions to visual concepts and back. In: CVPR (2015)
18. Firat, O., Sankaran, B., Al-Onaizan, Y., Vural, F.T.Y., Cho, K.: Zero-resource translation with multi-lingual neural machine translation. In: EMNLP (2016)
19. Gu, J., Cai, J., Joty, S., Niu, L., Wang, G.: Look, imagine and match: Improving textual-visual cross-modal retrieval with generative models. In: CVPR (2018)
20. Gu, J., Cai, J., Wang, G., Chen, T.: Stack-captioning: Coarse-to-fine learning for image captioning. In: AAAI (2018)
21. Gu, J., Wang, G., Cai, J., Chen, T.: An empirical study of language cnn for image captioning. In: ICCV (2017)
22. Gu, J., Wang, Z., Kuen, J., Ma, L., Shahroudy, A., Shuai, B., Liu, T., Wang, X., Wang, G., Cai, J., et al.: Recent advances in convolutional neural networks. Pattern Recognition (2017)

23. He, K., Zhang, X., Ren, S., Sun, J.: Deep residual learning for image recognition. In: CVPR (2016)
24. Hitschler, J., Schamoni, S., Riezler, S.: Multimodal pivots for image caption translation. In: ACL (2016)
25. Jean, S., Cho, K., Memisevic, R., Bengio, Y.: On using very large target vocabulary for neural machine translation. In: ACL (2015)
26. Jia, X., Gavves, E., Fernando, B., Tuytelaars, T.: Guiding long-short term memory for image caption generation. ICCV (2015)
27. Johnson, M., Schuster, M., Le, Q.V., Krikun, M., Wu, Y., Chen, Z., Thorat, N., Viégas, F., Wattenberg, M., Corrado, G., et al.: Google's multilingual neural machine translation system: enabling zero-shot translation. TACL (2016)
28. Kalchbrenner, N., Blunsom, P.: Recurrent continuous translation models. In: EMNLP (2013)
29. Karpathy, A., Fei-Fei, L.: Deep visual-semantic alignments for generating image descriptions. In: CVPR (2015)
30. Kingma, D., Ba, J.: Adam: A method for stochastic optimization. In: ICLR (2015)
31. Kulkarni, G., Premraj, V., Dhar, S., Li, S., Choi, Y., Berg, A.C., Berg, T.L.: Baby talk: Understanding and generating image descriptions. In: CVPR (2011)
32. Lin, T.Y., Maire, M., Belongie, S., Hays, J., Perona, P., Ramanan, D., Dollár, P., Zitnick, C.L.: Microsoft coco: Common objects in context. In: ECCV (2014)
33. Liu, C., Sun, F., Wang, C., Wang, F., Yuille, A.: Mat: A multimodal attentive translator for image captioning. In: IJCAI (2017)
34. Liu, S., Zhu, Z., Ye, N., Guadarrama, S., Murphy, K.: Improved image captioning via policy gradient optimization of spider. In: ICCV (2017)
35. Luong, M.T., Le, Q.V., Sutskever, I., Vinyals, O., Kaiser, L.: Multi-task sequence to sequence learning. In: ICLR (2016)
36. Luong, M.T., Sutskever, I., Le, Q.V., Vinyals, O., Zaremba, W.: Addressing the rare word problem in neural machine translation. In: ACL (2015)
37. Mao, J., Xu, W., Yang, Y., Wang, J., Huang, Z., Yuille, A.: Deep captioning with multimodal recurrent neural networks (m-rnn). In: ICLR (2014)
38. Mi, H., Sankaran, B., Wang, Z., Ittycheriah, A.: Coverage embedding models for neural machine translation. In: EMNLP (2016)
39. Papineni, K., Roukos, S., Ward, T., Zhu, W.J.: Bleu: a method for automatic evaluation of machine translation. In: ACL (2002)
40. Plummer, B.A., Wang, L., Cervantes, C.M., Caicedo, J.C., Hockenmaier, J., Lazebnik, S.: Flickr30k entities: Collecting region-to-phrase correspondences for richer image-to-sentence models. In: ICCV (2015)
41. Ranzato, M., Chopra, S., Auli, M., Zaremba, W.: Sequence level training with recurrent neural networks. In: ICLR (2016)
42. Rennie, S.J., Marcheret, E., Mroueh, Y., Ross, J., Goel, V.: Self-critical sequence training for image captioning. In: CVPR (2017)
43. Sutskever, I., Vinyals, O., Le, Q.V.: Sequence to sequence learning with neural networks. In: NIPS (2014)
44. Tu, Z., Lu, Z., Liu, Y., Liu, X., Li, H.: Modeling coverage for neural machine translation. In: ACL (2016)
45. Utiyama, M., Isahara, H.: A comparison of pivot methods for phrase-based statistical machine translation. In: NAACL (2007)
46. Vedantam, R., Lawrence Zitnick, C., Parikh, D.: Cider: Consensus-based image description evaluation. In: CVPR (2015)
47. Vinyals, O., Toshev, A., Bengio, S., Erhan, D.: Show and tell: A neural image caption generator. In: CVPR (2015)

48. Vinyals, O., Toshev, A., Bengio, S., Erhan, D.: Show and tell: Lessons learned from the 2015 mscoco image captioning challenge. PAMI (2016)
49. Vinyals, O., Toshev, A., Bengio, S., Erhan, D.: Show and tell: Lessons learned from the 2015 mscoco image captioning challenge. PAMI (2017)
50. Wu, H., Wang, H.: Pivot language approach for phrase-based statistical machine translation. Machine Translation (2007)
51. Wu, J., Zheng, H., Zhao, B., Li, Y., Yan, B., Liang, R., Wang, W., Zhou, S., Lin, G., Fu, Y., et al.: Ai challenger: A large-scale dataset for going deeper in image understanding. arXiv preprint arXiv:1711.06475 (2017)
52. Wu, Q., Shen, C., Liu, L., Dick, A., van den Hengel, A.: What value do explicit high level concepts have in vision to language problems? In: CVPR (2016)
53. Xu, K., Ba, J., Kiros, R., Cho, K., Courville, A., Salakhutdinov, R., Zemel, R.S., Bengio, Y.: Show, attend and tell: Neural image caption generation with visual attention. In: ICML (2015)
54. Yang, X., Zhang, H., Cai, J.: Shuffle-then-assemble: Learning object-agnostic visual relationship features. In: ECCV (2018)
55. Yao, T., Pan, Y., Li, Y., Qiu, Z., Mei, T.: Boosting image captioning with attributes. In: ICCV (2017)
56. You, Q., Jin, H., Wang, Z., Fang, C., Luo, J.: Image captioning with semantic attention. In: CVPR (2016)
57. Zahabi, S.T., Bakhshaei, S., Khadivi, S.: Using context vectors in improving a machine translation system with bridge language. In: ACL (2013)
58. Zhu, Y., Lu, S., Zheng, L., Guo, J., Zhang, W., Wang, J., Yu, Y.: Texygen: A benchmarking platform for text generation models. In: SIGIR (2018)
59. Zoph, B., Yuret, D., May, J., Knight, K.: Transfer learning for low-resource neural machine translation. In: EMNLP (2016)